\documentclass[sigconf]{acmart}

\copyrightyear{2021}
\acmYear{2021}
\setcopyright{rightsretained}
\acmConference[CIKM '21]{Proceedings of the 30th ACM International Conference on Information and Knowledge Management}{November 1--5, 2021}{Virtual Event, QLD, Australia}
\acmBooktitle{Proceedings of the 30th ACM International Conference on Information and Knowledge Management (CIKM '21), November 1--5, 2021, Virtual Event, QLD, Australia}
\acmDOI{10.1145/3459637.3482012}
\acmISBN{978-1-4503-8446-9/21/11}

\AtBeginDocument{%
  \providecommand\BibTeX{{%
    \normalfont B\kern-0.5em{\scshape i\kern-0.25em b}\kern-0.8em\TeX}}}

\usepackage{multirow} 
\usepackage[normalem]{ulem}
\usepackage{balance}

\begin{document}
\fancyhead{}
\title[A Portuguese-English Dataset for Question-Answering about the Ocean]{\textit{Pirá}: A Bilingual Portuguese-English Dataset \\ for Question-Answering about the Ocean}


\author{André F. A. Paschoal}
\authornote{Both authors contributed equally to this research.}
\orcid{}
\affiliation{%
  \institution{Escola de Artes, Ciências e Humanidades\\ Universidade de São Paulo}
  \streetaddress{}
  \city{}
  \country{}}
\email{andre.faleiros.paschoal@usp.br}

\author{Paulo Pirozelli}
\authornotemark[1]
\authornote{Corresponding author: paulo.pirozelli.silva@usp.br.}
\affiliation{%
  \institution{Instituto de Estudos Avançados \\ Universidade de São Paulo}
  \country{}
}
\email{paulo.pirozelli.silva@usp.br}

\author{Valdinei Freire, Karina V. Delgado}
\author{Sarajane M. Peres}
\orcid{}
\affiliation{%
  \institution{Escola de Artes, Ciências e Humanidades \\ Universidade de São Paulo}
  \streetaddress{}
  \city{}
  \country{}}
\email{{valdinei.freire,kvd,sarajane}@usp.br}

\author{Marcos M. José, Flávio Nakasato}
\author{André S. Oliveira, Anarosa A. F. Brandão} 
\author{Anna H. R. Costa, Fabio G. Cozman}
\affiliation{%
  \institution{Escola Politécnica \\ Universidade de São Paulo}
  \city{}
  \country{}
}
\email{{marcos.jose, flavio.cacao, andre.seidel}@usp.br}
\email{{anarosa.brandao,anna.reali,fgcozman}@usp.br}

\renewcommand{\shortauthors}{Paschoal and Pirozelli, et al.}

\begin{abstract}
Current research in natural language processing is highly dependent on carefully produced corpora. Most existing resources focus on English; some resources focus on languages such as Chinese and French; few resources deal with more than one language. This paper presents the \textit{Pirá} dataset, a large set of questions and answers about the ocean and the Brazilian coast both in Portuguese and English. \textit{Pirá} is, to the best of our knowledge, the first QA dataset with supporting texts in Portuguese, and, perhaps more importantly, the first bilingual QA dataset that includes this language. The \textit{Pirá} dataset consists of 2261 properly curated question/answer (QA) sets in both languages. The QA sets were manually created based on two corpora: abstracts related to the Brazilian coast and excerpts of United Nation reports about the ocean. The QA sets were validated in a peer-review process with the dataset contributors. We discuss some of the advantages as well as limitations of \textit{Pirá}, as this new resource can support a set of tasks in NLP such as question-answering, information retrieval, and machine translation.
\end{abstract}



\maketitle


\section{Introduction}
\label{sec:intro}

The best current solutions to question answering and reading comprehension tasks rely
on large scale datasets. 
That poses a problem for many languages; even though a number of datasets are available in English   \cite{rajpurkar2016squad, rajpurkar2018know, nguyen2016ms, clark2018think}, resources in other languages are rather scarce. 
While a few other languages have received some attention, such as Chinese \cite{CUI18.32, he2017dureader}, French \cite{FQuAD}, and German \cite{moller2021germanquad}, and while one can find multilingual datasets around \cite{lewis2019mlqa, liu2019xqa, xqaud}, many languages, such as Portuguese, still lag behind.
Question answering (QA) in non-English languages suffers from an additional difficulty: many, and in some cases most, of the documents used to answer questions are only available in English. Those working with non-English QA must then resort to automated translations without the support of curated bilingual datasets.

In this paper, we describe the creation of the \textit{Pirá} dataset, a high-quality question answering dataset for Portuguese and English that focuses on the ocean and the Brazilian coast.
\textit{Pirá} \footnote{The word \textit{Pirá}  means ``fish'' in Tupi-Guarani, a family of indigenous languages from South America that heavily influenced Brazilian Portuguese.} is an openly available bilingual scientific dataset built with the help of 254 
volunteer undergraduate and graduate students. To the best of our knowledge, the \textit{Pirá} dataset is the first QA dataset with supporting texts in Portuguese; more importantly, it is the first bilingual QA dataset where Portuguese is one of the languages. \textit{Pirá} is also the first QA dataset in Portuguese with unanswerable questions so as to allow the study of \textit{answer triggering}; finally, it is the first QA dataset 
that deals with scientific knowledge about 
the ocean, climate change, and marine biodiversity.


\textbf{Contributions.} We offer the following key contributions: \begin{enumerate}
    \item A bilingual (Portuguese-English) QA dataset about ocean data, biodiversity, and climate change, consisting of 4074 
    texts and 2261 
    QA sets. In our dataset, a {\it QA set} consists of four elements: a question in Portuguese and in English, and an answer in Portuguese and in English. 
    \item Methods both for enriching datasets through the production of equivalent answers and paraphrased questions, and  for manually assessing and describing QA datasets.
\end{enumerate}
Besides, we describe in this paper an open application for producing QA datasets 
based on supporting documents that can be used for crowdsourcing. 

The paper is structured as follows: In Section \ref{sec:background} we outline existing datasets and highlight the \textit{Pirá} dataset main features. In Section \ref{sec:method} we describe the protocol for building the dataset as well as the method for creating and evaluating questions. Section \ref{sec:datasets} explains the process for augmenting the dataset with the manual creation of answers and questions; it also describes three versions of the \textit{Pirá} dataset that we make available. In Section \ref{sec:analysis}, we present a preliminary analysis of the dataset, describing the main results obtained in the assessment step. Section \ref{sec:discussion} is dedicated to use cases of the \textit{Pirá} dataset, as well as to a discussion of some of its limitations. We then conclude the paper in Section \ref{sec:conclusion} with plans for future work. 


\section{Background and Motivation} 
\label{sec:background}

In this section we first summarize a few facts about the domain of our QA dataset,
namely, the ocean and in particular the Brazilian coast. This is a suitable domain
not only due to its intrinsic importance and complexity, but also because it naturally
lends itself to a dataset in Portuguese and English --- one of our main goals is
to understand the challenges that bilingual conversational agents may face.
We also present in this section a brief survey of existing resources for Natural
Language Processing in Portuguese.


\subsection{Domain} 
\label{subject}

More than 70 per cent of the surface of the planet is covered by the ocean and 95 per cent of  Earth's biosphere lies in it (adopting current view that there is a single connected ocean). Several economic activities directly depend on the ocean, such as fishing, tourism, and   extraction of natural resources. Currently, more than 80\% of the international trade is made by shipping. The ocean and its ecosystems also provide significant benefits to the global community, including climate regulation, coastal protection, food, employment, recreation, cultural well-being and spiritual bonds.

Recent change in weather patterns caused primarily by human-induced global warming are raising the ocean's temperature and threatening maritime ecosystems. The development of infrastructure in coastal areas, overfishing and garbage dumping are putting many species in danger. Anthropogenic noise is also disturbing maritime life \cite{un2017world, un2021world}.

All these factors demand a greater social awareness of the ocean fundamental importance to human life and the planet. For that reason, the United Nations (UN) established as two of its Sustainable Development Goals ``to conserve and sustainably use the oceans, seas and marine resources''   and ``take urgent action to combat climate change and its impacts''  \cite{un2015transforming}.
The ocean is studied in many fields, such as geology, oceanography, biology, and economics. Despite its importance, up to now no public dataset deals with these topics or, as to our knowledge, any close themes. By filling this gap, we hope that \textit{Pirá} dataset can stimulate more AI researchers to contribute with the advance and diffusion of knowledge on the sustainable use of the ocean.

\subsection{Existing resources for Portuguese} 
\label{language}

Compared to English, and even to other languages such as Chinese or German, resources in Portuguese are rather limited. Most existing resources are geared towards basic syntactic and semantic analysis, such as Mac-Morpho \cite{aluisio2003account, fonseca2013mac} for part-of-speech tagging, or PropBank~\cite{duran2012propbank} for Semantic role labeling. Amongst task-oriented resources and associated benchmarks we can cite ASSIN \cite{fonseca2016assin} for semantic similarity and textual entailment; SIMPLEX-PB \cite{hartmann2018simplex, hartmann2020simplex} for lexical simplification; and IDPT for irony detection \cite{de2014pathways, da2018detecccao, marten2021construction}.

QA datasets are particularly rare in Portuguese. Large multilingual QA datasets, for example, tend to ignore Portuguese \cite{asai2020xor, lewis2019mlqa, xqaud, clark2020tydi, liu2019xqa}, even though it is the sixth most spoken language in the world, with 221 million native speakers.\footnote{\url{https://en.wikipedia.org/wiki/List_of_languages_by_number_of_native_speakers}} Existing QA datasets in Portuguese usually consist of automatic translations of datasets in English, which are then sampled for manual editing. We mention, for instance, versions of SQuAD\footnote{\url{https://drive.google.com/file/d/1Q0IaIlv2h2BC468MwUFmUST0EyN7gNkn/view}} and GLUE\footnote{\url{https://drive.google.com/drive/folders/1WXdarHqxlJqKm30uD2LubwuLfpAGYZX0}} in Portuguese. The two exceptions are the ENEM-Challeng \cite{ENEM-Challenge} and MilkQA \cite{criscuolo2017milkqa}. The ENEM-Challenge is based on ENEM, the entrance examination valid for almost all universities in Brazil. The dataset contains 1,800 multiple choice questions on Humanities, Languages, Sciences and Mathematics, manually annotated with the types of background knowledge that are required to answer the questions. MilkQA consists of consumer questions from Embrapa's (Brazilian Agricultural Research Corporation) Dairy Cattle unity. The MilkQA dataset contains 2,657 anonymized pairs of questions and answers created directly in Portuguese, in which questions are associated with a pool of 50 candidate answers where only one answer is correct.






\section{Method} 
\label{sec:method}

\begin{figure*}[tp]
  \centering
  \includegraphics[width=0.85\textwidth]{ 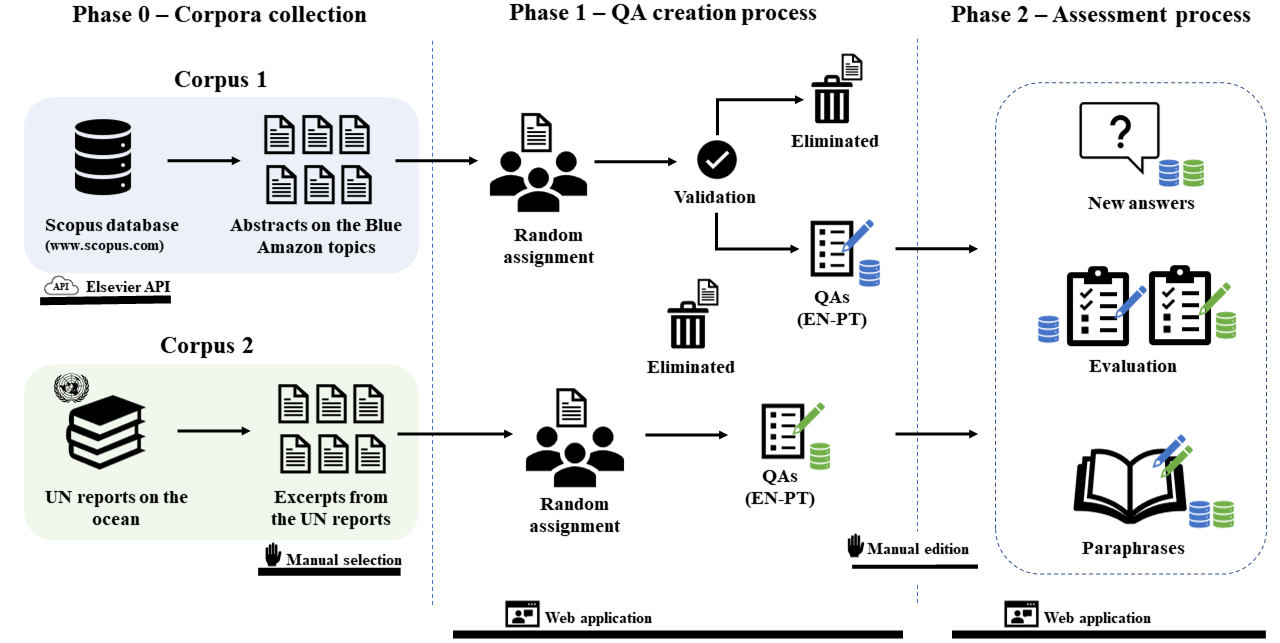}
  \caption{Overview of the \textit{Pirá} dataset generation process}
  \Description{}
  \label{overview}
\end{figure*}

The dataset generation process is depicted in Figure~\ref{overview}. Firstly, we collected two different corpora: abstracts of scientific papers about the Brazilian coast (also known as ``Blue Amazon''), and small excerpts of two books about the ocean organized by the United Nations. Secondly, QA sets were manually generated by a set of volunteers. The volunteers were undergraduate and graduate students and researchers at the University of São Paulo. The high educational level of participants let us build a scientific dataset with questions and answers that go beyond mere trivia.
In this second phase, volunteers received random texts from both corpora and had to create QA sets based on them. Participants were instructed to produce questions that could be answered with the use of the texts and no other source of information. Third, QA sets went through an extensive manual assessment process. Volunteers received QA sets produced in the previous phase by other individuals, and for each of these QA sets, they had to: i) answer the question in both languages without having access to the original answer; ii) assess the whole original QA set (the questions and respective answers) according to a number of aspects; and iii) paraphrase the original question. 

A total of 254 
volunteers took part in the activity (18 
researchers, 169 
undergraduate students, 67 
graduate students).\footnote{All individual-level information has been removed from the public dataset.} The remainder of this section describes each phase of the method in detail.

\subsection{Phase 0 - Corpora collection}

Two sets of texts were used as supporting documents for the QA sets generation.
Similarly to the approach taken by PubMedQA \citep{jin2019pubmedqa}, our Corpus 1  contains abstracts of scientific papers on the Brazilian coast topics. Abstracts were gathered from Elsevier's Scopus database,\footnote{\url{www.scopus.com}} an abstract and citation database that covers thousands of scientific journals, conference proceedings and books in different fields of knowledge. The construction of the corpus required us to filter relevant texts from the thousands of documents in the Scopus database.  First, we manually analyzed the results for several different keyword sets, in order to find accurate filtering queries and to minimize the number of false positives. Then an expert in the field evaluated the results of our queries and suggested changes to our set of keywords. Finally, we run a final search with the improved query 
and downloaded the abstracts together with their metadata. 


Corpus 2 consists of excerpts of two reports about the ocean organized by the United Nations, the \textit{World Ocean Assessment I} \cite{un2017world} and the \textit{World Ocean Assessment II} \cite{un2021world}. The excerpts were manually selected for the task, following some guidelines: they presented  relatively independent contents  and dealt with topics that could be understood by readers from the exact sciences with only a generic knowledge of other areas.

The abstracts contained in Corpus 1 are considerably more technical when compared to the text excerpts contained in Corpus 2. Table \ref{corpora} summarizes the main characteristics of the two corpora.

\begin{table} [tp]
  \caption{Summary of the textual corpora}
  \label{corpora}
  \begin{tabular}{lll}
    \toprule
    Characteristic & Corpus 1 & Corpus 2\\
    \midrule
    Subject & Brazilian coast & Ocean \\ 
    Type & Abstracts & Text excerpts \\
    Source & Scopus & UN reports  \\
    Number of texts & {3891} & {183}  \\ 
    Average size {\footnotesize (words)} & {201.72} & {344.88}\\ 
    Smallest document {\footnotesize (words)} & {15} & {98}\\ 
    Largest document {\footnotesize (words)} & {1176} &{1208}\\ 
    \bottomrule
  \end{tabular}
\end{table}

\subsection{Phase 1 - QA creation}

After collecting the two corpora, the next step focused on the construction of QA sets by volunteers. A web application was developed for that purpose. Because the corpora were distinct, the method had to be slightly adapted for each case. For this reason, we describe both procedures separately. 

\subsubsection{Corpus 1 - Scientific abstracts on Brazilian coast topics}

Because abstracts were automatically selected, some retrieved abstracts were in fact not related to our domain of interest. Such false positives should not be considered as a basis for constructing the QA dataset. To fix that, we carried out a validation step: after receiving a new abstract, volunteers were asked to confirm its adherence to the topic; in case the document was not related to our domain, participants were instructed to eliminate it. Those abstracts were then marked as false positives in our database and not distributed again in the whole process. If a volunteer thought a document actually dealt with the domain of interest, she was instructed to mark the text as a true positive. To help volunteers in the validation step, the web application provided them with a non-exhaustive list of topics related to the Brazilian coast.


After the validation step,   QA sets were generated. Volunteers were asked to produce up to three question/answer pairs per selected abstract.
As our dataset is bilingual, volunteers were asked to also produce translations of the questions and answers (recall that each QA set contains four elements: a question in Portuguese, an answer in Portuguese, a question in English, and an answer in English). Volunteers were allowed to use automatic translation tools as long as they checked the translations. They should not use the internet, however, to search for other sources of information about the subject discussed in the abstract under analysis.

Before they started generating QA sets, volunteers were advised as follows: questions should not be generic, but rather they should be contextualized according to the subject discussed in the abstract under analysis; answers should be based on the abstract, but they did not need to be a span, or segment of text, of the abstract; answers should be short, but not repeating the question again; yes or no questions should be avoided (in order to obtain more complex QA sets); and the volunteers should try to diversify the types of questions, varying the use of question words, such as ``why'', ``when'', ``how'', and so on. 

We created an optimal scoring scheme (15 points) to encourage participation and reduce early dropout. In fact, the score served as a guide to indicate what we considered to be a minimum expected engagement. Although volunteers were not required to achieve this number of points, the scheme acted as a way to show them whether they were falling short of our expectations and/or contributing beyond them. The application also had a scoring system that rewarded a higher number of questions for abstract: one point for one question, three points for two questions and five points for three questions. The aim was to get more questions per abstract, thus obtaining more complex and diversified questions, instead of easy, and repeatable, patterns of question. Table~\ref{resultsQAcreation} shows the results of this process on Corpus 1.


\subsubsection{Corpus 2 - Text excerpts from UN reports about the ocean}


For Corpus 2, the validation step was not necessary as texts came from books that specifically dealt with the subject. 
In this case, volunteers were asked to make up two QA sets for text excerpt. We lowered the number of questions based on the difficulty volunteers voiced regarding Corpus 1. Again, each QA set comprised four elements: question in Portuguese, answer in Portuguese, question in English, and answer in English. The remaining instructions were the same as for Corpus 1. 

We also created a scoring scheme (8 points) to reward volunteers' engagement with the activity. Volunteers had to create two QA sets in order to move to another excerpt; they received one point for QA set submitted. Table~\ref{resultsQAcreation} shows the results from the QA set generation process regarding Corpus 2.

\begin{table} [tp]
  \caption{Results of the QA generation process}
  \label{resultsQAcreation}
  \begin{tabular}{p{4.4cm}ll}
    \toprule
    Characteristics & Corpus 1 & Corpus 2 \\
    \midrule
    \#Abstracts/excerpt texts used & {532} & {149}\\ 
    \#Abstracts eliminated & {727} & - - \\ 
    \#QA sets generated & {1675} & {586}\\ 
    Texts per volunteer & {2.96} & {2.01}\\ 
    QA sets per volunteer & {9.31} & {7.92}\\ 
    QA sets per text & {3.15} & {3.94}\\ 
   {Words per question (avg.) (EN | PT)} & {13.37 | 14.07} & {12.11 | 13.06}\\ 
    {Words per answer (avg.) (EN | PT)} & {13.34 | 14.56} & {14.90 | 16.23}\\ 
    \bottomrule
  \end{tabular}
\end{table}


\subsubsection{Manual editing}

The volunteers are well educated people, most of them familiar with scientific texts. Nevertheless, the complexity of the topics, the academic style of the texts, and the fact that individuals were not native English speakers led to a number of mistakes. To overcome that, the QA sets produced in Phase 1 were manually edited for spelling and grammar. 


We also took the opportunity to remove questions that were too broad, such as ``Which process does this paper illustrate?''. We did not eliminate, however, questions that just did not contextualize it enough, since that aspect should be evaluated in the second phase of the QA dataset generation process. Table \ref{questions} presents some examples of the final state of the questions and answers.

\begin{table*} [tp]
  \caption{Examples of questions and answers}
  \label{questions}
  \begin{tabular}{lp{7cm}p{7cm}}
    \toprule
     & Context & QA set\\
    \midrule
    Corpus 1 & {\small The greenhouse effect and resulting increase in the Earth's temperature may accelerate the mean sea-level rise. The natural response of bays and estuaries to this rise, such as this case study of Santos Bay (Brazil), will include change in shoreline position, land flooding and wetlands impacts. The main impacts of this scenario were studied in a physical model built in the Coastal and Harbour Division of Hydraulic Laboratory, University of São Paulo, and the main conclusions are presented in this paper. [...]} & {\small \textbf{Question-EN:} What are some consequences of the greenhouse effect in bays and estuaries?} \newline {\small \textbf{Question-PT:} Quais são algumas consequências do efeito estufa em baías e estuários?} \newline {\small \textbf{Answer-EN:} Change in shoreline position, land flooding and wetlands impacts.} \newline {\small \textbf{Answer-PT:} Mudança na posição da linha costeira, inundações de terra e impactos nas zonas húmidas.} \\ 
    
    Corpus 2 & {\small [...] Even in the open ocean, climate warming will increase ocean stratification in some broad areas, reduce primary production and/or result in a shift in productivity to smaller species (from diatoms of 2-200 microns to picoplankton of 0.2-2 microns) of phytoplankton. This has the effect of changing the efficiency of the transfer of energy to other parts of the food web, causing biotic changes over major regions of the open ocean, such as the equatorial Pacific.} & {\small \textbf{Question-EN:} What is one of the effects of the increased ocean stratification, resulting from the climate warming?} \newline {\small \textbf{Question-PT:} Qual é um dos efeitos do aumento na estratificação dos oceanos, resultante do aquecimento do clima?} \newline {\small \textbf{Answer-EN:} Changing the efficiency of energy transfer within the food web in the ocean.} \newline {\small \textbf{Answer-PT:} Alteração da eficiência da transferência de energia dentro da teia alimentar no oceano.} \\ 
    
    \bottomrule
  \end{tabular}
\end{table*}

\subsection{Phase 2 - Assessment process} 



After  Phase 1 was finished, volunteers were invited to evaluate the QA sets in a number of ways. The abstracts and excerpts, with the corresponding QA sets, were randomly assigned to volunteers in order to check their quality. We ensured that volunteers would not evaluate QA sets that they themselves created. QA sets were treated individually in this process; i.e., a volunteer evaluated a single QA set for a given text, even if in the previous phase more than one QA set had been made for that same text. The assessment procedure consisted of three steps described in the remainder of this section. 

\subsubsection{Answering}

In the first step, volunteers received an abstract or excerpt and a question associated with it (both in English and in Portuguese). They then had to answer it in both languages, taking into account only the information available via the abstract/excerpt text. The volunteers could also pass to another abstract/excerpt text in case they were not able to answer the current question. 

\subsubsection{Assessment}
\label{sec:assess}

In the second step, volunteers were presented again to the same abstract/excerpt text and question, but now also with the \textit{original} answer and the new answer elaborated in the previous step. They were then asked to evaluate this whole set according to the following aspects: 

\begin{itemize}
    \item answer two yes/no questions: {(1) \textit{Is this a generic question? Example: ``How big is the oil field?''}; (2) \textit{Can you answer the question only with the information provided by the text?}}
    \item evaluate three statements about the QA set according to a Likert scale (1 - Strongly disagree, 2 - Disagree, 3 - Neither agree nor disagree, 4 - Agree, 5 - Strongly agree): {(1) \textit{The question makes sense;} (2) \textit{The question is hard;} (3) \textit{Your answer and the original answer are equivalent.}}
    \item indicate the type of question: {\textit{who, what, where, when, why, how, none of them}}.
\end{itemize}

\subsubsection{Rewriting}

One of the main shortcomings of QA datasets based on texts is that questions and answers often share words with the texts, a semantic relationship that can be easily grasped by reading comprehension models. For that reason, even relatively simple models based on word matching heuristics (such as counting the number of words that are repeated in the question and candidate sentences) can obtain performances comparable to state-of-the-art baselines \cite{weissenborn2017making}. 
Given this point, some datasets try to fix this dependency \textit{a posteriori}, through a debiasing procedure, excluding QA sets that are easily answerable in this way \citep{sakaguchi2020winogrande}. Instead, we established an additional step in the assessment phase in which volunteers were asked to paraphrase the question if they considered it acceptable. That procedure resembles the one used in SelQA \citep{jurczyk2016selqa}, in which the annotators were instructed to paraphrase the questions to reduce the co-occurrence of words between the texts, and the questions and answers. In our dataset, however, the rewriting process was executed by a different individual, bringing an additional factor of diversity to the generation of QA sets.\footnote{In fact, the process of asking volunteers to avoid using the same words in questions as they appear in the texts (or of writing and then rewriting them, as in SelQA \cite{jurczyk2016selqa}) was already among the instructions of the first phase in our method.} Some examples of paraphrases are shown in Table \ref{paraphrase}.

\begin{table*} [htp]
  \caption{Examples of paraphrases}
  \label{paraphrase}
  \begin{tabular}{lp{7cm}p{7cm}}
    \toprule
     & Old question & New question\\
    \midrule
    Corpus 1 & {\small \textbf{Question-EN:} What impact did the Guanabara oil spilling pollution caused in foraminiferal taphonomic assemblages?} \newline {\small \textbf{Question-PT:} Qual o impacto que a poluição do derramamento do óleo da Guanabara causou nas assembléias tafonômicas foraminíferas?} & {\small \textbf{Question-EN:} What effect did the pollution from the Guanabara oil disaster have on foraminiferal taphonomic assemblages?} \newline {\small \textbf{Question-PT:} Qual o efeito da poluição do desastre do óleo da Guanabara sobre as assembléias tafonômicas foraminíferas?}\\ 
    
    
    Corpus 2 & {\small \textbf{Question-EN:} What are the consequences on the plankton ecosystem as ocean water warms?} \newline {\small \textbf{Question-PT:} Quais as consequências no ecossistema de plânctons à medida que a água do oceano aquece?} & {\small \textbf{Question-EN:} What are the consequences of the rising temperature of ocean water for the plankton ecosystem?} \newline {\small \textbf{Question-PT:} O que acontece com o ecossistema do plâncton quando a temperatura do oceano sobe?}\\ 
    
    \bottomrule
  \end{tabular}
\end{table*}


\section{Dataset augmentation and versions}
\label{sec:datasets}

With the new questions and answers produced during Phase 2, we were able to expand and enrich the original dataset, obtaining more lexically diverse QA sets. Furthermore, the assessments carried out by the volunteers allowed us to qualify the QA sets. Three versions of the dataset were created: \textit{Pirá-F} (filtered), \textit{Pirá-T} (triggering), and \textit{Pirá-C} (complete). 


\textit{Pirá-F} is the filtered version of the dataset. It includes only QA sets that volunteers indicated as not generic, that could be answered with only the information provided by the text, and moreover that volunteers agreed (4) or strongly agreed (5) that made sense. When volunteers reported in Phase 2 that their answer did not coincide with the original one (i.e., 1-2 in the Likert scale for ``Your answer and the original answer are equivalent''), we excluded the new answer from the dataset. Finally, we also included the paraphrases. The dataset contains up to four possible combinations of QA sets: original question, original answer; original question, new answer; paraphrase, original answer; paraphrase, new answer.

\textit{Pirá-T} is the answer-triggering version of the dataset. It includes all QA sets generated in Phases 1 and 2 (meaningful and meaningless, contextualized and non-contextualized, answered by the text only or not). As for \textit{Parí-F}, we also excluded
questions that were assessed with one or two points for the statement ``Your answer and the original answer are equivalent''. In addition, the answers (original e new) for questions that volunteers regarded meaningless (1-2 in the Likert scale for ``The question makes sense'') were filled with `` '' (blank value). 
In datasets that every question possess an answer, models can always guess an answer (and sometimes be right just by luck). Instead, by allowing some questions not to have an answer, models are required to learn when they are able to answer a given question, a task known as ``answer triggering'' (which is getting more  important in QA datasets \cite{yang-etal-2015-wikiqa, jurczyk2016selqa, rajpurkar2018know}).

For the sake of completeness, we also make available \textit{Pirá-C}, with all the questions and answers produced in the generation and assessment phases.

\section{Dataset analysis} 
\label{sec:analysis}

In this section, we describe the results obtained in Phase 2. In general, volunteers that were assigned to Corpus 1 (Brazilian coast) took more time to finish the activity, as compared to those working with Corpus 2 (Ocean). This result was expected, given the specialized nature of the abstracts, in comparison to the texts extracted from the UN reports, which are intended for a larger audience. In fact, we also selected the text excerpts that seemed more convenient to readers in the composition of Corpus 2. Table~\ref{resultsQAvalidation} shows the results of the QA validation process on Corpus 1 and 2. Almost the whole dataset was validated, resulting in the generation of 1876 paraphrases. 

\begin{table} [tp]
  \caption{Results of the QA validation process}
  \label{resultsQAvalidation}
  \begin{tabular}{p{4.1cm}ll}
    \toprule
    Characteristics & Corpus 1 & Corpus 2 \\
    \midrule
    QA sets validated & {1506} & {556}\\ 
    
    QA sets validated with paraphrase & {1342} & {534}\\ 
    
    Percentage of QA sets validated & {89.91\%} & {94.88\%}\\ 
    
    {Words per paraphrase question (avg.) (EN | PT)} & {13.70 |  14.56} & {12.34 | 13.10}\\

    {Words per new answer (avg.) (EN | PT)} & {12.45 | 13.74} & {14.10 | 15.45}\\ 
    
    \bottomrule
  \end{tabular}
\end{table}

\begin{figure}[tp]
  \centering
  \includegraphics[width=0.49\textwidth]{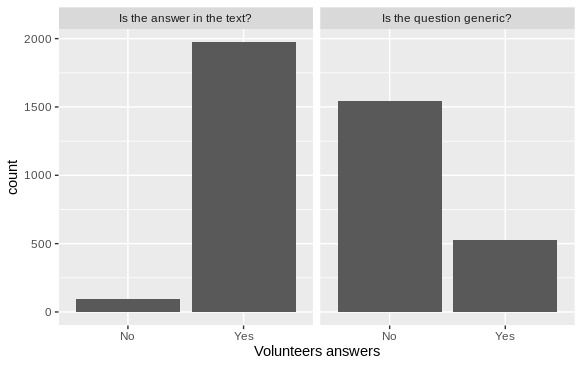}
  \caption{Answers for the Yes/No questions in Phase 2.}
  \Description{}
  \label{yes-no}
\end{figure}

\begin{figure}[tp]
  \centering
  \includegraphics[width=0.49\textwidth]{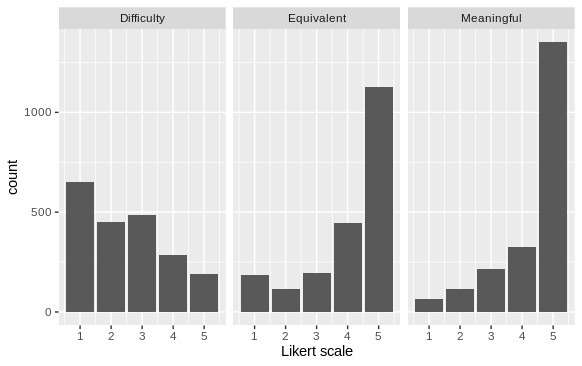}
  \caption{Answers for Likert-scale questions in Phase 2 (cf. Section~\ref{sec:assess})} 
  \Description{}
  \label{likert}
\end{figure}

Figure~\ref{yes-no} shows the distributions of answers to Yes/No questions for the whole dataset. Almost all questions were considered as being answerable by the content of the text only (95.2\%). Approximately three quarters of the questions were properly contextualized, according to volunteers (74.8\%). Figure~\ref{likert} shows the results for the three Likert-scale assessments of QA sets' features. Volunteers generally disagreed that questions were difficult (1 - 31.5\%, 2 - 22\%). Most of the times, questions were considered meaningful (5 - 65.3\%, 4 - 15.8\%) and the original and new answers as equivalent (5 - 54.6\%, 4 - 21.7\%).

\begin{figure}[tp]
  \centering
  \includegraphics[width=0.49\textwidth]{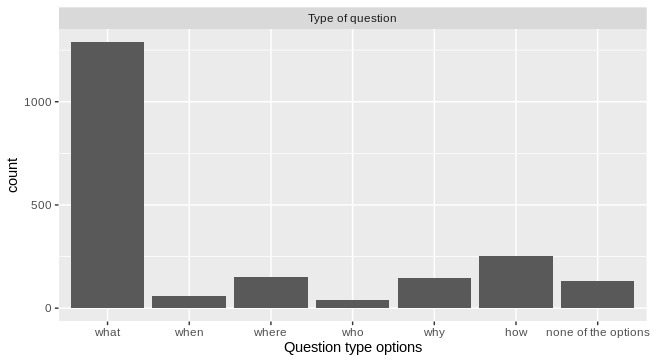}
  \caption{Answers for type of questions in Phase 2.}
  \Description{}
  \label{type}
\end{figure}

Volunteers also classified questions as regards their type. ``What'' questions made up the majority (62.1\%), followed by ``How'' (12\%), ``Where'' (7.4\%), and ``Why'' (6.9\%). ``When'' questions represented 2.8\% of the total and ``Who'' questions 1.9\%. In 4.7\% of the cases, questions were not classified in any of the categories. Results can be seen in Figure~\ref{type}.

In general, assessments were similar for both corpora. A few differences should be stressed, though. For example, answers in Corpus 1 were slightly more generic, according to volunteers (26.8\% and 20.6\%, respectively). 
Proportionally, volunteers working with Corpus 2  more often stated they ``totally agreed'' the questions made sense (70.1\%, against 63.6\% on Corpus 1). More people responded they ``neither agree nor disagree'' the question was difficult for Corpus 1 than for Corpus 2 (22\% and 26.6\%, respectively). As regards the equivalence between the original and the new answer, results were 
lower for Corpus 1 --- 9.7\% of the volunteers ``totally disagreed'' the answers were equivalent for questions based on texts from Corpus 1, whereas only 5.9\% said the same for questions based on texts from Corpus 2. Finally, there was a higher incidence of ``where'' questions (8.6\% for Corpus 1 and 4.1\% for Corpus 2) in Corpus 1 and of ``how'' questions in Corpus 2 (10.7\% for Corpus 1 and 15.6\% for Corpus 2). 



\begin{figure*}[tp]
  \centering
  \includegraphics[width=0.5\textwidth]{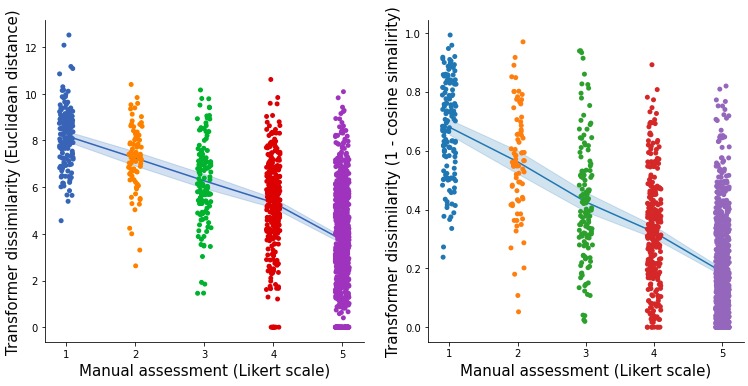}
  \caption{Manual \textit{vs} transformer assessment of sentence equivalence (Corpus 1)}
  \Description{}
  \label{equivalence_A}
\end{figure*}
\begin{figure*}[tp]
  \centering
  \includegraphics[width=0.475\textwidth]{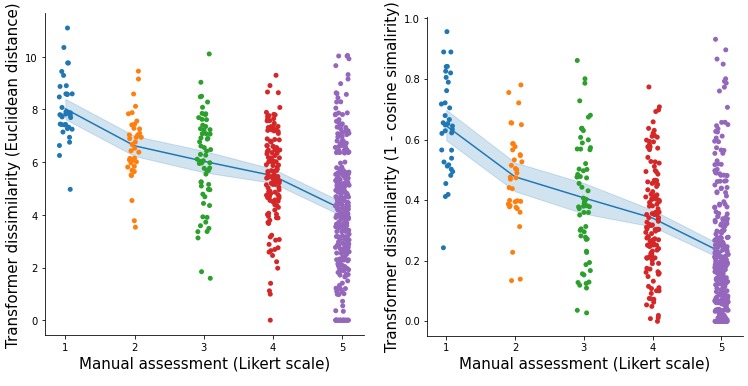}
  \caption{Manual \textit{vs} transformer assessment of sentence equivalence (Corpus 2)}
  \Description{}
  \label{equivalence_B}
\end{figure*}

We then tested the capacity of pre-trained Transformers \cite{vaswani2017attention} sentence embeddings to determine equivalences in English sentences, as compared to human judgments. The embeddings were generated by Sentence-BERT \cite{reimers2019sentence}, the state-of-the-art model for semantic textual similarity assignment.\footnote{We applied ``paraphrase-distilroberta-base-v1'' checkpoint from Hugging Face, a flavor of Sentence-BERT on top of DistilBERT model, fine-tuned on several paraphrase-related tasks.} Dissimilarity between answers was measured using Euclidean distance and cosine similarity. We then compared these results with the manual assessments of equivalence generated in Phase 2. The results are reported in Figures \ref{equivalence_A} and \ref{equivalence_B}. In average, answers to which volunteers attributed a higher degree of equivalence achieved higher similarity scores through Sentence-BERT. Nonetheless, there are some divergences between embedding and human-assigned equivalences, as it can be seen on both figures. 

We then carefully examined some of the answers in which manual and automatic assessments diverged the most. To do that, we selected the answers in Corpus 2 which received 5 (cf. Likert scale) in the assessment phase as regards to sentence equivalence but received the highest scores for Euclidean distance (in the Transformer analysis). Three main factors explained the divergence between manual and transformer sentence assessments. First, some volunteers gave short answers while others gave long ones. Second, answers with the same content often had relevant lexical and syntactic differences. Finally, some answers differed in the presence of mathematical and logical reasoning over the text. Table~\ref{similarity} brings examples of each situation. 

\begin{table*} [htp]
  \caption{Human \textit{vs} Transformers embeddings semantic similarity}
  \label{similarity}
  \begin{tabular}{p{3.4cm}p{5.6cm}p{6.8cm}}
    \toprule
    Divergence & Question & Answers\\
    \midrule
    Short/long answer & {\small What is the proportion of women among the seafarers?} & {\small \textbf{Original answer:} Worldwide approximately two per cent of the seafarers are women.} \newline {\small \textbf{New answer:} About two per cent.}\\ 
    Lexical/syntactic diversity & {\small What are the risks that nanoplastics and microplastics in the oceans impose to the human health?} & {\small \textbf{Original answer:} For now, according to the European Food Safety Authority, it is not possible to evaluate the risks.} \newline {\small \textbf{New answer:} There's not enough scientific evidence to back the claim that micro and nano plastics pose a risk to human health.}\\ 
    
    Reasoning & {\small How much was the increase in per capita consumption of fish from 2013 to 2016?} & {\small \textbf{Original answer:} 0.8 kg.} \newline {\small \textbf{New answer:} Per capita food fish consumption was estimated at 20.3 kg in 2016, compared with 19.5 kg in 2013.}\\ 
    \bottomrule
  \end{tabular}
\end{table*}

\section{Discussion} 
\label{sec:discussion}

In this section, we discuss   possible applications, limitations of the \textit{Pirá} dataset, and lessons learned from creating the dataset. 

\subsection{Impact and applications} 
\label{sec:impact}


The \textit{Pirá} dataset is a valuable resource for reading comprehension and information retrieval. It is particularly useful for researchers working with bilingual tasks, since it mimics a common problem faced by them: that of accessing texts in English to answer questions in their own language. Four characteristics make \textit{Pirá} dataset particularly interesting for that matter: i) it is bilingual; ii) questions were based on corpora with different complexities; iii) QA sets were manually assessed in a number of ways; and iv) the dataset was enriched by paraphrases. 

A first use case for the \textit{Pirá} dataset is in information retrieval. The task, in that case, is finding, for a question, the text which contains the answer to it. Neural retrievers, such as the Dense Passage Retriever, for instance, are trained using QA datasets \cite{karpukhin2020dense}.

The most important use case for the \textit{Pirá} dataset, however, is connected with reading comprehension. Within this complex task, at least two specific paths can be followed, which we refer to as generative \cite{lewis2021retrievalaugmented} and extractive \cite{rajpurkar2016squad} answering. In the generative type, an end-to-end (i.e., question-to-answer) system searches for relevant passages in a corpus and generates answers from them. 
This task is the one most naturally adapted to the \textit{Pirá} dataset, given the mostly non-span structure of its answers and the high-quality manual assessments made on it.

The \textit{Pirá} dataset can also be employed for extractive answering, in which given a question and a text, the task is that of finding the corresponding span that contains the answer. Although the answers in \textit{Pirá} dataset are not necessarily spans, the dataset can be approximated in this way, as long as one keeps in mind that results tend to be not as good as in a purely extractive dataset.

Some aspects of the \textit{Pirá} dataset make it particularly challenging for these tasks. The first reason is connected to the scientific nature of its corpora, with questions that can be hard even for people  to answer. Furthermore, the fact volunteers had to answer and paraphrase each other's questions contributes to a diversified dataset. This is fundamental in reducing biases derived from annotation artifacts, which are frequent in crowdsourcing efforts \cite{gururangan2018annotation}.  

Finally, other possible applications for the \textit{Pirá} dataset are classification tasks (taking into account the keywords for Corpus 1 and book sections for Corpus 2 to establish data labels) and for machine translation fine-tuning. 

As for the different versions of the dataset, the goal is to  test  distinct aspects of models. \textit{Pirá-F} is the filtered version of the dataset, composed of easier QA sets.
As for \textit{Pirá-T}, the idea is to foster answer triggering tasks, so that models cannot simply ``guess'' an answer, but must also consider whether a question can actually be answered. Finally, \textit{Pirá-C} is the complete dataset, which can be filtered according to the researchers' interests.

\subsection{Limitations} 
\label{sec:limitations}

\textbf{Contextualization.} The main limitation of the \textit{Pirá} dataset certainly refers to issues of generality and lack of contextualization. Even seemingly well-defined questions can carry assumptions that go almost unnoticed; especially due to the lack of clear time frames (e.g., ``How much will Petrobras invest for gas production in the state of Bahia [in 2010]?''). This is particularly taxing given that some data in the QA sets are regularly updated: for instance, the number of deep-water oil wells on the Brazilian coast.

\textbf{Translation.} Another difficulty is related to typical problems of translation. Many of the expressions in these specialized texts do not possess  widely accepted translations. Several alternatives are found in the dataset to deal with these expressions: keeping them in the original form (which sometimes uses words that have already been incorporated into Portuguese, such as ``design'', and other less popular ones), translating them literally (sometimes through adaptations that do not adjust perfectly with Portuguese, such as ``stressors'' $\rightarrow$ ``\textit{estressores}''), and even translating only part of the expression (''blocky effect'' $\rightarrow$ ''\textit{efeito} blocky'' or `` offshore fields'' $\rightarrow$ ``\textit{campos} offshore''). Similarly, stylistic difficulties are encountered: for example, in which cases should the name of a site, a technique or a project be translated? Ideally, the solution should depend on a convention. In the dataset, some volunteers left expressions in English, others preferred to translate them and still others placed the original and its corresponding translation together. The same happened with Brazilian companies and placenames: should ``Marlin Sul'' be kept unchanged or be translated to ``South Marlin''? In addition, English has much more flexibility to create compounds than Portuguese, and constructions that sound natural in the former may become clumsy in the latter: ``How are visual feeding species affected by low-oxygen water areas?'' $\rightarrow$ ``\textit{Como as áreas de água com baixo oxigênio afetam as espécies que usam a visão para se alimentar}?''. 

\textbf{Linguistic idiosyncrasies.} Other difficulties are related to essential differences between Portuguese and English. A very common example derives from the fact that Portuguese has genders (male and female). Although for most words there is a unique translation, in many cases the common Portuguese speaker finds it difficult to decide which gender to use. Particularly challenging are the nominal phrases that involve elliptical constructions.  
``Why has Cengroup Petroleum signed a contract with the Azerbaijani government?'' $\rightarrow$ ``\textit{Por que \textbf{A} Cengroup Petroleum assinou contrato com o governo do Azerbaijão}?'', due to the word ``\textit{empresa}'' [company -- female in Portuguese]  or ``\textit{Por que \textbf{O} Cengroup Petroleum assinou contrato com o governo do Azerbaijão}?'', due to the word ``\textit{grupo}'' [group -- male in Portuguese] or ``\textit{petróleo}'' [petroleum -- male in Portuguese]?

\textbf{Portuguese-bias.} Because QA sets were produced by Portuguese-speaking people, there is a tendency for questions in the dataset to be better formulated in Portuguese than in English. This ``Portuguese-bias'', although not the ideal scenario, may perhaps counterbalance the preponderance of English datasets or English-adapted datasets.

\textbf{Structure.} Even if correct, several questions are hard to grasp. For instance: ``Are Petrobras semisubmersible platforms SS-20 used in connection with aut-leg and conventional mooring systems  operating at which water depth?''. One  reason for this kind of question may be a difficulty in handling complex nominal phrases, leading volunteers to begin questions with these expressions. In addition, many  abstracts in Corpus 1 have been written by non-English speakers in academic language, making it even more difficult to understand their specialized content.

\textbf{Method subdetermination.} Despite giving detailed instruction and explaining the activity in online meetings, rules were not sufficient to completely standardize results. For example, although volunteers were supposed to write short answers, the results varied a great deal. Moreover, despite asking volunteers to avoid ``yes/no questions'', some of them can be found in the dataset. Other questions were simply not addressed by our instructions: for instance, some volunteers wrote questions with several sentences, a possibility that we did not explictly forbid.




\subsection{Further steps}

Although preliminary tasks have been tried with \textit{Pirá}, we intend to conduct more extensive and systematic explorations. A first step would be to produce robust baselines for three tasks: information retrieval, extractive answering, and generative answering. Those tasks can be then tested with different input-output language combinations as follows: 

\begin{itemize}
    \item Information retrieval: Portuguese (question only); English (question only);
    \item Extractive answering: Portuguese $\rightarrow$ English; English $\rightarrow$ English; Any language $\rightarrow$ English;
    \item Generative answering: Portuguese $\rightarrow$ Portuguese; English $\rightarrow$ English; Any language $\rightarrow$ Same language as the input.
\end{itemize}


We also plan to expand \textit{Pirá} by regularly asking for volunteers, thus including other topics related to the ocean and the Brazilian coast. 
We would like also to expand the size of the dataset by filtering natural questions related to the ocean topics from the datasets Natural Questions \cite{kwiatkowski2019natural} and MS-Marco \cite{nguyen2016ms},
and to perform experiments with dialogues about the texts of Corpus 2 \cite{choi2018quac}.

Another line of investigation is the comparison of the results in the three tasks mentioned above by considering both languages, Portuguese and English, so as to establish a baseline for other studies in these languages,
and possibly for bilingual tasks in general. We would like to conduct further analyzes on the variations arising from human behavior during the generation and assessment of QAs sets (e.g., sentence length; syntactic complexity, as measured by dependency tree; and topics). 


Finally, we hope to examine more in-depth the results on equivalence assessments, in order to improve estimates of semantic similarity for open answers.


\section{Final Remarks} 
\label{sec:conclusion}




In this paper, we presented \textit{Pirá}, a bilingual QA  dataset containing 2261 
manually generated QA sets about the Brazilian coast and the ocean. Three versions of the dataset are available: \textit{Pirá-F}, containing meaningful and contextualized QA sets; \textit{Pirá-T}, including questions without answers; and \textit{Pirá-C}, which brings the complete dataset. 
We hope these datasets provide valuable training data for bilingual QA systems; careful manual assessments for questions and answers; and data enrichment based on paraphrases. Also, we presented a replicable method for dataset generation (with freely available code for an open source application). Possible applications of the dataset are: reading comprehension, information retrieval, and machine translation. Finally, we hope that \textit{Pirá} can contribute to AI research as applied to the ocean sciences, climate change, and marine biodiversity.


\textbf{Availability.} All versions of the dataset and the corresponding code are freely available for download and use under the \textit{Creative Commons Attribution 4.0 International license}. \textit{Pirá} is published as an online resource at Github.\footnote{https://github.com/C4AI/Pira} 
Instructions and code for the web application are also available at Github.

\textbf{Ethical issues.} This project fits into Art. 1, VI, of the Resolution Nº 510 of the Brazilian Ministry of Health, which dispenses the necessity of submitting it to an ethical committee. Before engaging in the activity, volunteers signed a consent form stating they were voluntarily participating of the activity and also giving up the rights on the produced material. Furthermore, no information of personal level is included in the dataset.

\begin{acks}
The authors are grateful for the collaboration of Dr. Eduardo Aoun Tannuri (Escola Politécnica - Universidade de São Paulo) for his valuable contributions regarding   domain knowledge.

This work was carried out at the Center for Artificial Intelligence (C4AI-USP), with support by the São Paulo Research Foundation (FAPESP grant \#2019/07665-4) and by the IBM Corporation. This research was also partially supported by Itaú Unibanco S.A.; M. M. José, F. Nakasato and A. S. Oliveira have been supported by the Itaú Scholarship Program (PBI) of the Data Science Center (C2D) of the Escola Politécnica da Universidade de São Paulo. A. H. R. Costa and F. G. Cozman thank the support of the National Council for Scientific and Technological Development of Brazil (CNPq grants \#310085/2020-9 and \#312180/2018-7, respectively).
\end{acks}

\bibliographystyle{ACM-Reference-Format}
\balance
\bibliography{sample-base}


\end{document}